# Emotional voice conversion: Theory, databases and ESD

Kun Zhou [a,*], Berrak Sisman [b], Rui Liu [b], Haizhou Li [a,c]

[a] Department of Electrical and Computer Engineering, National University of Singapore, Singapore
[b] Information Systems Technology and Design Pillar, Singapore University of Technology and Design, Singapore
[c] Machine Listening Lab, University of Bremen, Germany

ABSTRACT

In this paper, we first provide a review of the state-of-the-art emotional voice conversion research, and the existing emotional speech databases. We then motivate the development of a novel emotional speech database (*ESD*) that addresses the increasing research need. With this paper, the *ESD* database[1] is now made available to the research community. The *ESD* database consists of 350 parallel utterances spoken by 10 native English and 10 native Chinese speakers and covers 5 emotion categories (neutral, happy, angry, sad and surprise). More than 29 h of speech data were recorded in a controlled acoustic environment. The database is suitable for multi-speaker and cross-lingual emotional voice conversion studies. As case studies, we implement several state-of-the-art emotional voice conversion systems on the *ESD* database. This paper provides a reference study on *ESD* in conjunction with its release.

## 1. Introduction

Speech is more than lexical words and carries the emotion of the speakers. Previous research (Mehrabian and Wiener, 1967) indicates that the spoken words, which are referred to as verbal attitude, convey only 7% of the message when communicating feelings and attitudes. Non-verbal vocal attributes (38%) and facial expression (55%) have a major impact on the expression of social attitudes. Non-verbal vocal attributes reflect the emotional state of a speaker and play an important role in daily communication (Arnold, 1960). Emotional voice conversion (EVC) is a technique that aims to convert the emotional state of the utterance from one to another while preserving the linguistic information and speaker identity, as shown in Fig. 1(a). It allows us to project the desired emotion into a human voice, for example, to act or to disguise one's emotions. Speech synthesis technology has been progressing by leaps and bounds towards human-like naturalness. However, it has not adequately equipped with human-like emotions (Schuller and Schuller, 2018). Therefore, emotional voice conversion bears huge application potential in human–machine interaction (Tits et al., 2020; Pittermann et al., 2010; Crumpton and Bethel, 2016; Li et al., 2019), for example, to enable robots to respond to humans with emotional intelligence.

In general, voice conversion aims to convert the speaker identity in human speech while preserving its linguistic content, which is also referred to as speaker voice conversion (Childers et al., 1989), as shown in Fig. 1(b). Earlier studies of voice conversion are focused on modeling the mapping between source and target features with some statistical methods, which include Gaussian mixture model (GMM) (Toda et al., 2007), partial least square regression (Helander et al., 2010), frequency warping (Erro et al., 2009a) and sparse representation (Çişman et al., 2017; Sisman et al., 2018b, 2019b). Deep learning approaches, such as deep neural network (DNN) (Chen et al., 2014; Lorenzo-Trueba et al., 2018a), recurrent neural network (RNN) (Nakashika et al., 2014), generative adversarial network (GAN) (Sisman et al., 2018c) and sequence-to-sequence model with attention mechanism (Zhang et al., 2019b; Huang et al., 2019) have advanced the state-of-the-art. For effective modeling, parallel training data are required in general. Recently, voice conversion techniques with non-parallel training data have been studied, for example, domain translation (Kaneko and Kameoka, 2017; Kameoka et al., 2018), multitask learning (Zhang et al., 2019a) and speaker disentanglement (Hsu et al., 2016, 2017; Tanaka et al., 2019) are among the successful attempts. The recent progress in voice conversion becomes the source of inspiration for emotional voice conversion studies.

Both speaker voice conversion and emotional voice conversion aim to preserve the speech content and to convert para-linguistic information. In speaker voice conversion, the speaker identity is thought to be characterized by the speaker's physical attributes, which are determined by the voice quality of the individual. Thus, spectrum conversion has been the focus (Kain and Macon, 1998; Desai et al.,

* Corresponding author.
*E-mail addresses:* zhoukun@u.nus.edu (K. Zhou), berraksisman@u.nus.edu (B. Sisman), liu_rui@sutd.edu.sg (R. Liu), haizhou.li@nus.edu.sg (H. Li).

[1] https://hltsingapore.github.io/ESD/.

2010; Sisman, 2019). It is considered that prosody-related features are speaker-independent, that are to be carried forward from the source to the target (Liu et al., 2020a). However, in emotional voice conversion, emotion is inherently supra-segmental and complex that involves both spectrum and prosody (Arias et al., 2020; Schuller and Schuller, 2020). Therefore, it is insufficient to convert the emotion through frame-wise spectral mapping (Erro et al., 2009b). Segmental level prosodic dynamics needs to be properly dealt with (Tao et al., 2006).

The early studies on emotional voice conversion include GMM technique (Aihara et al., 2012; Kawanami et al., 2003), sparse representation technique (Aihara et al., 2014), or an incorporated framework of hidden Markov model (HMM), GMM and fundamental frequency (F0) segment selection method (Inanoglu and Young, 2009). The recent studies on deep learning have seen remarkable performance, such as DNN (Lorenzo-Trueba et al., 2018a; Luo et al., 2016, 2017), highway neural network (Shankar et al., 2019a), deep bi-directional long-short-term memory network (DBLSTM) (Ming et al., 2016a), and sequence-to-sequence model (Robinson et al., 2019; Le Moine et al., 2021). Beyond parallel training data, new techniques have been proposed to learn the translation between emotional domains with CycleGAN (Zhou et al., 2020b; Shankar et al., 2020b) and StarGAN (Rizos et al., 2020), to disentangle the emotional elements from speech with auto-encoders (Zhou et al., 2021b; Gao et al., 2019; Cao et al., 2020; Schnell et al., 2021), and to leverage text-to-speech (TTS) (Kim et al., 2020; Zhou et al., 2021a) or automatic speech recognition (ASR) (Liu et al., 2020b). Such framework generally works well in speaker-dependent tasks. Studies have also revealed that the emotions can be expressed through some universal principles that are shared across different individuals and cultures (Ekman, 1992; Kane et al., 2014; Manokara et al., 2020), that motivates the study of multi-speaker (Shankar et al., 2019, 2020; Liu et al., 2020b), and speaker-independent emotional voice conversion (Zhou et al., 2020c; Choi and Hahn, 2021).

Speaker-independent emotional voice conversion studies call for a large multi-speaker emotional speech database (Zhou et al., 2020c). However, existing databases, such as VCTK database (Yamagishi et al., 2019), CMU-Arctic database (Kominek and Black, 2004), and Voice Conversion Challenge (VCC) corpus (Toda et al., 2016; Lorenzo-Trueba et al., 2018b; Zhao et al., 2020), are designed for speaker voice conversion. The study of emotional voice conversion is facing the lack of a suitable database. To our best knowledge, there is no large-scale, multi-speaker, and open-source emotional speech database just yet. The lack of data greatly limits the emotional voice conversion research (Schuller et al., 2013). In addition, an overview of existing databases in Section 3 identifies that imperfect recording condition and limited lexical variability are other limitations. Based on the review of the current emotional voice conversion research landscape, we propose the development of a novel database to address the needs.

This paper is organized as follows. We first give a comprehensive overview of recent studies on emotional voice conversion in Section 2. We discuss the 19 existing databases (Zhou et al., 2021c) in Section 3. We then formulate the design of a novel *ESD* database for speaker-independent emotional voice conversion, that is also suitable for other speech synthesis tasks, such as mono-lingual or cross-lingual speaker voice conversion and emotional text-to-speech. The *ESD* database consists of a total of 29 h of audio recordings from 10 native English speakers and 10 native Chinese speakers, covering 5 different emotion categories (neutral, happy, angry, sad and surprise). It represents one of the largest emotional speech databases publicly available, in terms of speaker and lexical variability. All the recordings are conducted in the studio with professional devices to guarantee audio quality. In Section 5, we report the performance for the *ESD* database in emotional voice conversion. In Section 6, we show the promise of the *ESD* database on other voice conversion and text-to-speech tasks. Finally, Section 7 concludes the study.

## 2. Overview of emotional voice conversion

Emotional voice conversion is the enabling technology for many applications, such as expressive text-to-speech (Liu et al., 2020c; Tits et al., 2019a), speech emotion recognition (SER) (Bao et al., 2019; Rizos et al., 2020) and conversational agents (Tits et al., 2020). It has attracted much attention in recent years. Nearly one-third of the research papers under the prosody and emotion category at INTERSPEECH 2020 is about emotional voice conversion. As most of the state-of-the-art emotional voice conversion frameworks are based on data-driven techniques, they hugely depend on the training data (Inanoglu and Young, 2009; Zhou et al., 2020c; Elgaar et al., 2020). We start with an introduction to emotion in speech. Then, we give a comprehensive overview of emotional voice conversion studies according to the type of training data.

### 2.1. *Emotion in speech*

It is well known that emotion can be rendered in both the acoustic speech and spoken linguistic content (Schuller and Schuller, 2020). Psychological studies also show that acoustic attitudes, cf. emotional prosody, play a bigger role than spoken words in human–human communication (Mehrabian and Wiener, 1967). Since the emotional prosody is the complex interplay of multiple acoustic attributes, it is not straightforward to precisely characterize speech emotion (Hirschberg, 2004). When modeling the emotion, two research problems are generally considered: (1) how to describe and represent emotions; and (2) how to model the process of emotional expression and perception by humans.

#### 2.1.1. *Emotion representation*

Emotion can be characterized with either categorical (Whissell, 1989; Ekman, 1992) or dimensional representations (Russell, 1980; Schroder, 2006). With emotion-denoting labels, the emotion category approach is the most straightforward way to represent emotion. One of the most well-known categorical approaches is Ekman's six basic emotions theory (Ekman, 1992), where emotions are grouped into six discrete categories, i.e., angry, disgust, fear, happy, sad, and surprise, that are adopted in many emotional speech synthesis studies (Schröder, 2009; Barra-Chicote et al., 2010; Zhou et al., 2020c). However, such discrete representations do not seek to model the subtle differences between human emotions (Nose et al., 2007; Tao et al., 2006; Schuller and Schuller, 2018) so as to control the rendering of voice. Another way is to model the physical properties of emotion expression. One example is Russell's circumplex model (Russell, 1980; Soleymani et al., 2017), which is defined by arousal, valence, and dominance (Gunes and Schuller, 2013). For example, in the valence–arousal (V–A) representation (Xue et al., 2018), happy speech is characterized by positive valence and arousal values, while sad speech is characterized by all negative values. On the other hand, angry can be divided into hot and cold angry, which correspond to the full-blown angry and milder angry in psychology (Biassoni et al., 2016).

In general, both categorical and dimensional representations of emotions have been widely used in both emotion recognition (Schuller, 2018) and emotional voice conversion (Luo et al., 2019; Zhou et al., 2020b; Ming et al., 2015; Xue et al., 2018). The study on representation learning (Liu et al., 2020c) represents a new way of emotion representation, that further calls for large scale emotion-annotated speech data.

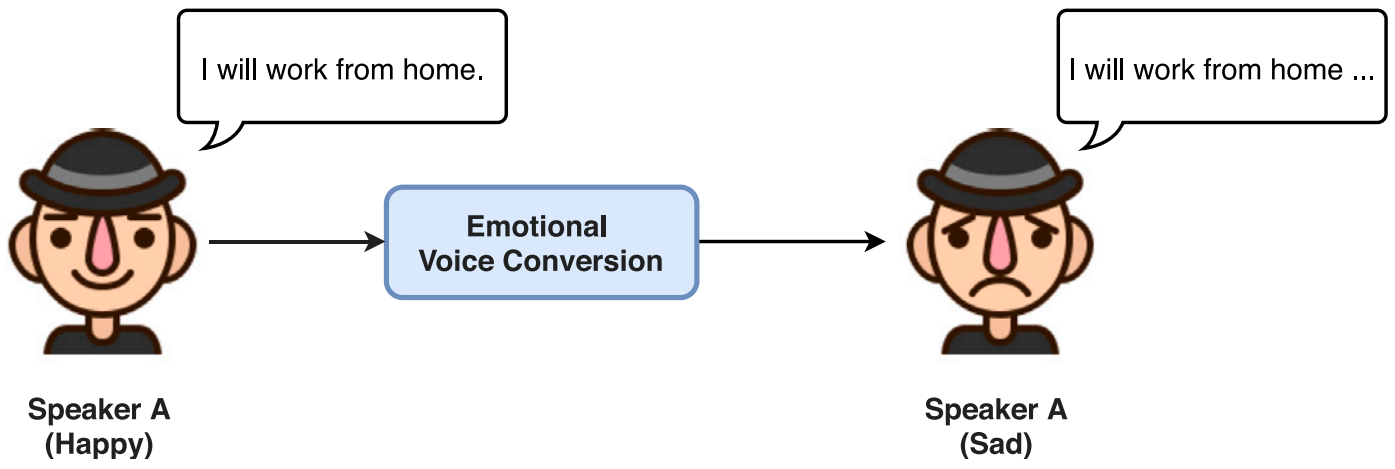

(a) A typical emotional voice conversion system is trained with speech data from the same speaker with different emotions. At run-time, the system converts the emotional state from one to another [45, 48].

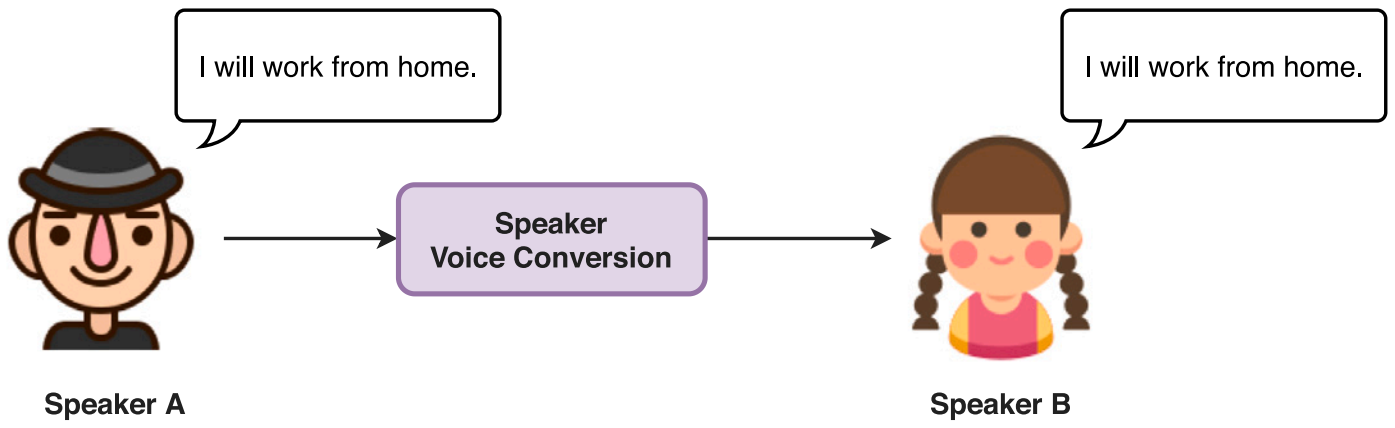

(b) A typical speaker voice conversion system is trained with speech data from different speakers. At run-time, the system converts the speaker identity from one to another [69, 18].

**Fig. 1.** A comparison between emotional voice conversion and speaker voice conversion.

#### 2.1.2. *Emotion perception and production*

Another research problem is how to model the process of emotional expression and perception. In Brunswik's model (Brunswik, 1956), it is suggested that the perception of emotion is multi-layered (Huang and Akagi, 2008). This model has been widely used in speech emotion recognition (Li and Akagi, 2016; Elbarougy and Akagi, 2014), where emotion categories, semantic primitives and acoustic features constitute the layers from top to bottom, respectively. Suppose that emotion production is an inverse process of emotion perception, an inverse three-layered model was studied for speech emotion production (Xue et al., 2018). The idea is further explored in emotional voice conversion (Xue et al., 2018).

Prosodic features concerning voice quality, speech rate, and fundamental frequency (F0), such as spectral features, duration, F0 contour and energy envelope are commonly used as the acoustic features in emotion-related studies (Schroder, 2006; Erickson, 2005). In emotional voice conversion, we are interested in transforming such acoustic features to render the emotions (Luo et al., 2016; An et al., 2017; Zhou et al., 2021b). In emotion recognition, we also rely on similar acoustic features (Schuller and Schuller, 2020; Latif et al., 2020), for example, expert-crafted features such as those extracted with openSMILE (Eyben et al., 2010, 2013), and learnt acoustic features from the spectrum (Ghosh et al., 2016). In emotional speech synthesis, both rule-based (Benesty et al., 2007; Zovato et al., 2004), and data-driven techniques with statistical modeling (Barra-Chicote et al., 2010; Yamagishi et al., 2005) or deep learning methods (An et al., 2017; Lorenzo-Trueba et al., 2018a) rely on speech databases for emotion analysis and production.

With the advent of deep learning, it is possible to describe different emotional styles in the continuous space with deep features learnt by neural networks (Zhou et al., 2021c; Schuller and Schuller, 2020; Tits et al., 2019b). Unlike human-crafted features, deep emotional features are less dependent on human knowledge (Schuller and Schuller, 2020), thus more suitable for emotional style transfer (Zhou et al., 2021c). Recently, deep emotional features are used to project an emotional style to a target utterance for emotional voice conversion (Zhou et al., 2021c). In this paper, we also visualize and evaluate the deep emotional features extracted from *ESD* database learnt by neural networks in Section 4.4. The studies on emotion analysis, recognition and synthesis have been inspiring emotional voice conversion studies.

### 2.2. *Emotional voice conversion with parallel data*

Early studies on emotional voice conversion mostly rely on parallel training data, i.e., a pair of utterances of the same content by the same speaker but with different emotions. During training, the conversion model learns a mapping from the source emotion A to the target emotion B through the paired feature vectors. In general, as shown in Fig. 2, an emotional voice conversion process typically consists of three steps, i.e. feature extraction, frame alignment and feature mapping, where dynamic time warping (DTW) (Müller, 2007) and model-based alignment with speech recognizer (Ye and Young, 2004) or attention mechanism (Zhang et al., 2019b; Tanaka et al., 2019) are commonly used for frame alignment.

#### 2.2.1. *Feature extraction and modeling*

Given the paired utterances, feature extraction seeks to obtain features that characterize emotion in speech. The resulting paired feature sequences are aligned to obtain frame-level alignment. It is common to use low-dimensional spectral representations from the high-dimensional spectra for modeling and manipulation. The commonly used spectral features include Mel-cepstral coefficients (MCC), linear predictive cepstral coefficients (LPCC), and line spectral frequencies (LSF).

As emotion is complex with multiple signal attributes concerning spectrum and prosody, both spectral and prosodic components need the same level of attention in emotional voice conversion. Several prosodic features are usually considered, such as pitch, energy, and duration. We note that F0 is an essential prosodic component that describes intonation, either linguistic or emotional, at different duration ranging from syllable to utterance (Veaux and Rodet, 2011). The methods to model the F0 variants include stylization methods (Teutenberg et al., 2008; Obin and Beliao, 2018) and multi-level modeling (Wang et al., 2017; Obin et al., 2011). As a multi-level modeling method, continuous wavelet transform (CWT) has been widely used to model hierarchical prosodic features, such as F0 (Suni et al., 2013; Ming et al., 2015; Luo et al., 2017) and energy contour (Şişman et al., 2017; Sisman and Li, 2018b; Sisman et al., 2019b). With CWT analysis, a signal can be decomposed into frequency components and represented with different temporal scales. CWT has shown to be effective for modeling the speech prosody (Ming et al., 2016b; Suni et al., 2013), and has been successfully applied into various emotional voice conversion frameworks such as exemplar-based (Ming et al., 2015) or other deep learning-based (Luo et al., 2019; Ming et al., 2016a; Zhou et al., 2020b) frameworks.

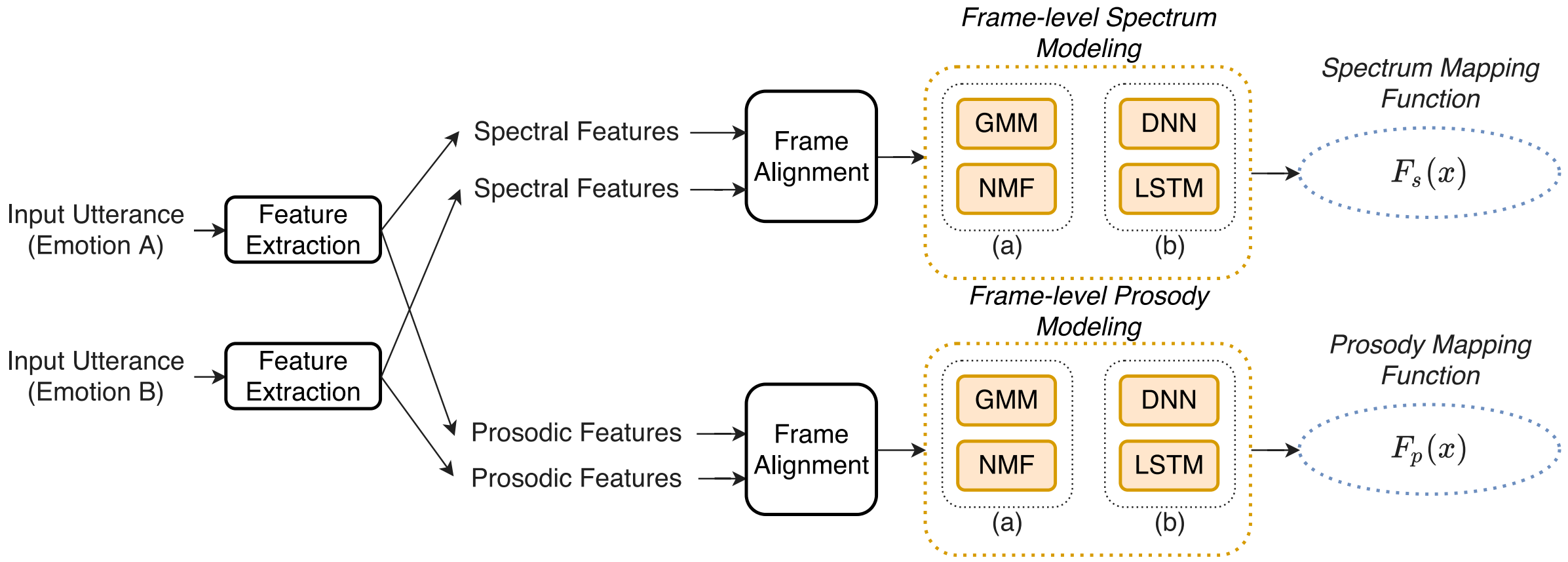

**Fig. 2.** A frame-level parallel emotional voice conversion framework with separate spectrum and prosody training using: (a) statistical methods, or (b) deep learning methods, to derive spectrum mapping function $F_s(x)$ and prosody mapping function $F_p(x)$. The yellow boxes represent the models involved in the training. The dotted box (a) includes examples of statistical approaches (Aihara et al., 2012, 2014), and (2) includes examples of deep learning approaches (Ming et al., 2015; Luo et al., 2016). (For interpretation of the references to color in this figure legend, the reader is referred to the web version of this article.)

### 2.2.2. Feature mapping

In a typical flow of emotional voice conversion, feature mapping aims to capture the relationship between the source and target features. With aligned paired utterances, feature mapping can be achieved by a statistical model (Inanoglu and Young, 2007). In Tao et al. (2006) and Wu et al. (2009), researchers propose to use a classification and regression tree to decompose the pitch contour of the source speech into a hierarchical structure, then followed by GMM and regression-based clustering methods. In Aihara et al. (2012), a GMM-based emotional voice conversion framework is proposed to learn the spectrum and prosody mapping. An exemplar-based emotional approach is introduced in Aihara et al. (2014), where parallel exemplars are used to encode the source speech signal and synthesize the target speech signal. The idea is further extended to a unified exemplar-based emotional voice conversion framework (Ming et al., 2015) that learns the mappings of spectral features and CWT-based F0 features simultaneously.

The neural network approach represents the state-of-the-art solution to feature mapping. DNN (Lorenzo-Trueba et al., 2018a), deep belief network (DBN) (Luo et al., 2016), highway neural network (Shankar et al., 2019a), and DBLSTM (Ming et al., 2016a) are examples that perform spectrum and prosody mapping with parallel training utterances. It is noted that frame-level feature mapping does not explicitly handle the mapping of duration, which is however an important element of prosody. The sequence-to-sequence encoder–decoder architecture represents a solution to duration mapping (Sutskever et al., 2014; Vaswani et al., 2017). With the attention mechanism, neural networks learn the feature mapping and alignment during training and automatically predict the output duration during run-time inference. The encoder–decoder model (Robinson et al., 2019) is an example, where pitch and duration are jointly modeled. In general, the more accurate alignment would help to build a better feature mapping function, that also explains why these frameworks are all built with parallel data.

## 2.3. Emotional voice conversion with non-parallel data

In practice, parallel speech data is expensive and difficult to collect. Thus, emotional voice conversion techniques with non-parallel data are more suitable for real-life applications (Zhou et al., 2020b, 2021b; Schuller and Schuller, 2020). We refer *non-parallel* data to the multi-emotion utterances, that do not share the same lexical content across emotions. Neural solutions have enabled many emotional voice conversion frameworks with non-parallel data. Next, we discuss two typical non-parallel methods, that are (1) auto-encoder (Kingma and Welling, 2013), and (2) CycleGAN (Zhu et al., 2017) methods. We summarize the ways to learn from non-parallel training data into three scenarios,

- Emotional domain translation,
- Disentanglement between emotional prosody and linguistic content,
- Leveraging TTS or ASR systems.

### 2.3.1. Emotional domain translation

Emotion voice conversion with non-parallel training data learns the unique emotion patterns and project them to the new utterances. To achieve this, translation models between the source and target domains have been studied, such as GANs (Emir Ak et al., 2019; Ak et al., 2020a, 2019, 2020b), CycleGANs (Zhu et al., 2017), Augmented CycleGAN (Almahairi et al., 2018) and StarGAN (Choi et al., 2018) in computer vision problems.

As shown in Fig. 3, CycleGAN is based on the concept of adversarial learning (Goodfellow et al., 2014), that is to train a generative model to find an optimal solution through a min–max game between the generator and the discriminator. In general, a CycleGAN is incorporated with three losses, that are (1) adversarial loss, (2) cycle-consistency loss, and (3) identity mapping loss. The adversarial loss measures how distinguishable between the converted and target feature distributions. The cycle-consistency loss eliminates the need for parallel training data through circular conversion, and the identity-mapping loss helps preserve the linguistic content between the input and output (Kaneko and Kameoka, 2017; Fang et al., 2018; Kaneko and Kameoka, 2018).

In emotional domain translation (Zhou et al., 2020b), CycleGANs are used to model the spectrum and prosody mapping between the source and target. For example, an adaptation of CycleGAN blended with implicit regularization is proposed (Shankar et al., 2020b) to predict the F0 contour, where the generators are composed of a trainable and a static component. These frameworks have shown the effectiveness to convert a pair of emotions.

Extending from two domains to multiple domains, StarGAN (Choi et al., 2018) is introduced to learn a multi-domain translation, that works for many-to-many speaker voice conversion (Kameoka et al., 2018; Kaneko et al., 2019). StarGAN is an improvement of CycleGAN, which consists of one generator/discriminator pair parameterized by class, and a domain classifier to verify the class correctness. In this way, StarGAN allows for multi-class to multi-class domain conversion. In Rizos et al. (2020), a StarGAN is adopted to learn spectral mappings between multiple emotional domains, and validated by data augmentation of emotion recognition. These emotional voice conversion studies represent the successful attempts to learn an emotion domain translation without the need for parallel data.

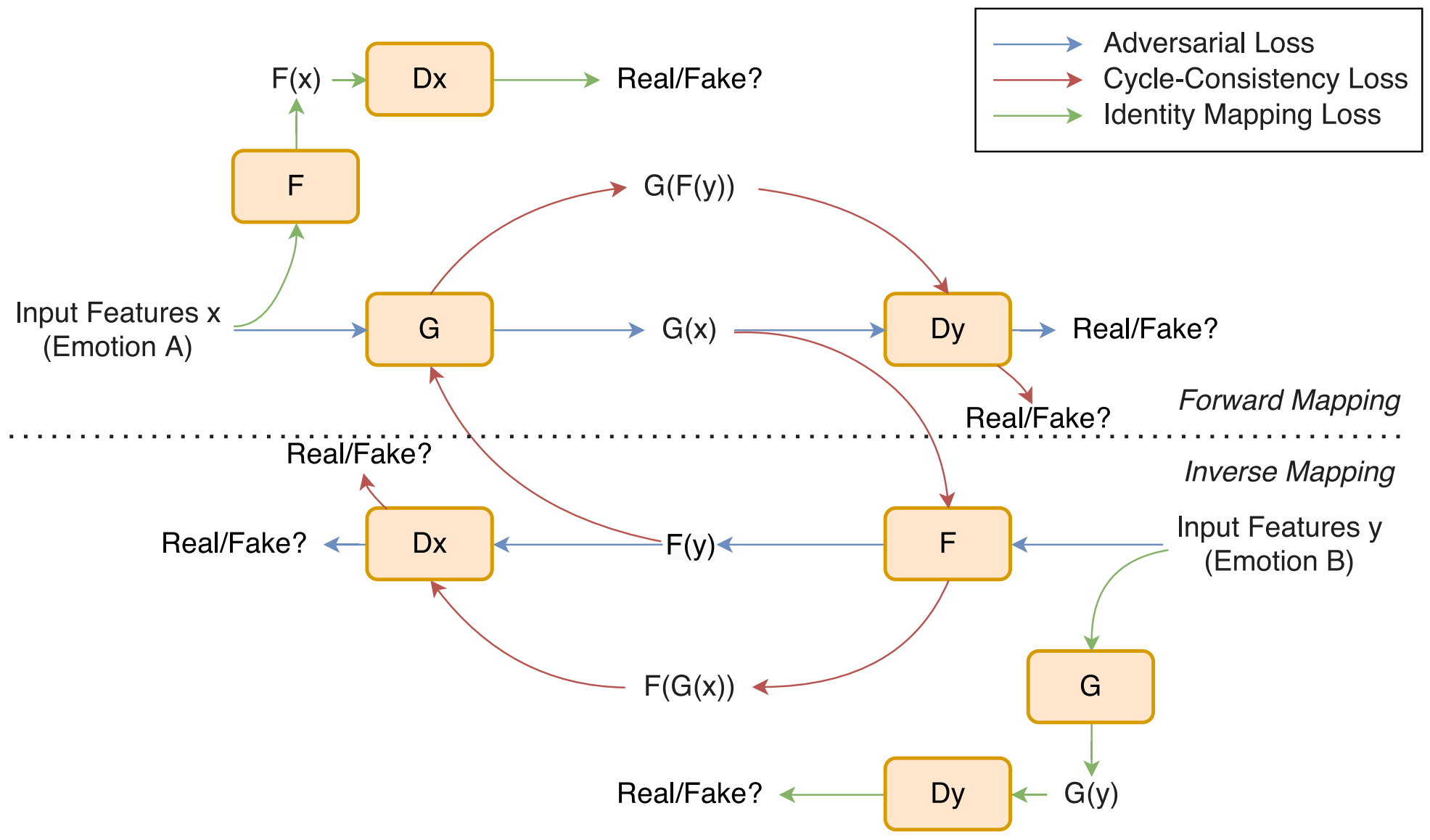

**Fig. 3.** A CycleGAN-based emotional voice conversion framework (Zhou et al., 2020b; Shankar et al., 2020b) learns both forward and inverse mappings between two different emotion domains through adversarial loss, cycle-consistency loss, and identity-mapping loss.

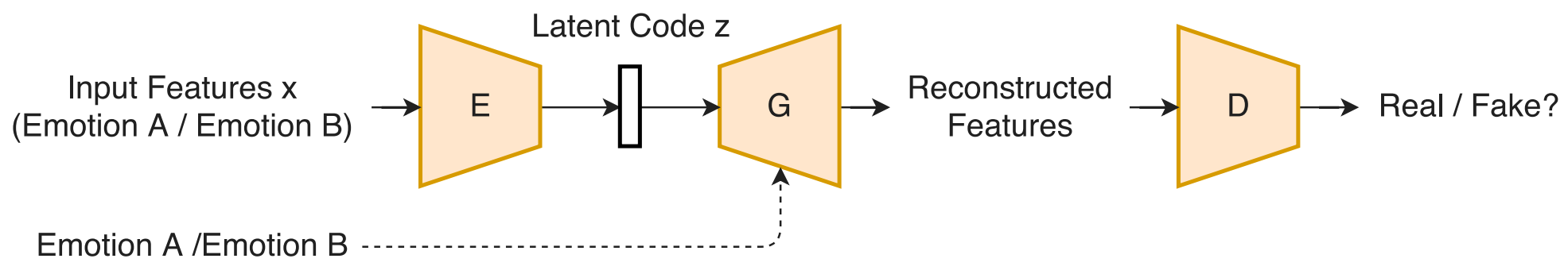

**Fig. 4.** An emotional voice conversion framework based on VAE-GAN (Cao et al., 2020) or VAW-GAN (Zhou et al., 2020c, 2021b,c) learns to disentangle emotional elements and recompose speech by assigning new emotional state. Emotion information (Emotion A/ B) is used as the condition of the generator, which can be an one-hot emotion identity (Zhou et al., 2020c, 2021b; Cao et al., 2020), or a deep feature representation for emotion (Zhou et al., 2021c).

#### *2.3.2. Disentanglement between emotional prosody and linguistic content*

In the context of emotional voice conversion, speech can be considered as a composition of linguistic content and speaking style. If we were able to disentangle the emotional style from the linguistic content, we could modify the emotional style independently without changing the linguistic content. Auto-encoder (Kingma and Welling, 2013) is one of the techniques which is commonly used for speech disentanglement and reconstruction. Studies have shown the effectiveness of auto-encoder (Hsu et al., 2016) and its variants (Hsu et al., 2017; Chou et al., 2019; Wu and Lee, 2020) in disentangling the speaker information from the content, thus they are widely used in speaker voice conversion (Sisman et al., 2020) and singing voice conversion (Lu et al., 2020).

As shown in Fig. 4, a typical auto-encoder-based emotional voice conversion mainly consists of three components, those are (1) encoder, (2) generator, and (3) discriminator. The encoder learns to generate a latent code $z$ frame-by-frame that is supposed to contain emotion-agnostic information such as linguistic content and speaker identity. The generator learns to reconstruct the input features from the latent code $z$ and the emotion identity, the discriminator learns to distinguish the reconstructed features with the real features. Auto-encoder is easy to train owing to its simple architecture (Elgaar et al., 2020; Cao et al., 2020; Gao et al., 2019).

In practice, we can train an emotion classifier to derive deep feature representation from speech, i.e., deep emotional feature. The generator is then conditioned on such deep emotional feature or one-hot emotion identity during the training. In this way, we learn the frame-level content-related representations from the speech in an unsupervised manner (Cao et al., 2020; Zhou et al., 2020c, 2021b). VAE-GAN (Cao et al., 2020) and VAW-GAN (Hsu et al., 2017; Zhou et al., 2020c) are successful attempts. Another idea is for an auto-encoder to learn an emotion-invariant latent code and an emotion-related style code in latent space (Gao et al., 2019). In doing so, a pair of content encoder and style encoder is used to disentangle the emotional style from the speaking content. In sum, auto-encoder is proven effective in learning disentangled representations, which makes training with non-parallel data possible.

#### *2.3.3. Leveraging TTS or ASR systems*

The voice conversion task shares many common attributes with the TTS task. For example, both require a high-quality speech vocoder. Studies also show that voice conversion benefits from the knowledge about linguistic content in the speech. For example, speaker voice conversion successfully leverages TTS (Zhang et al., 2019; Huang et al., 2019; Zhang et al., 2021) or ASR systems (Sisman et al., 2018a; Tian et al., 2019) that are phonetically-informed and trained on large speech corpus.

Inspired by the success in speaker voice conversion, multi-task learning between emotional voice conversion and text-to-speech is studied (Kim et al., 2020). In this framework, a single sequence-to-sequence model is trained to optimize both VC and TTS, in which the VC system benefits from the latent phonetic representation learnt by TTS during the training. At the run-time inference, given a reference speech, the system can transfer the emotional style of the input utterances from one to another. Experimental results also show that the VC and TTS system with multi-task learning outperforms a VC-alone system.

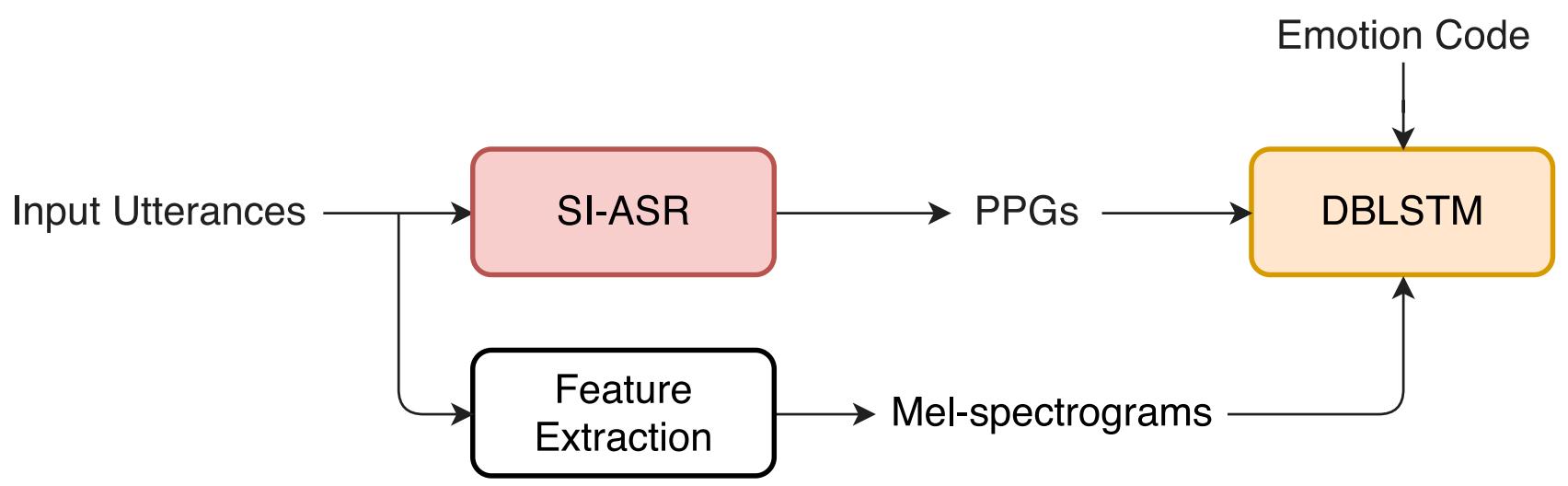

**Fig. 5.** An example of PPG-based emotional voice conversion framework that leverages from ASR (Liu et al., 2020b). A pre-trained speaker-independent automatic speech recognition (SI-ASR) is used to extract PPGs from the input utterances. Red box represents the model that is pre-trained. (For interpretation of the references to color in this figure legend, the reader is referred to the web version of this article.)

It is known that the prosody rendering of emotional speech consists of both speaker-dependent and speaker-independent, which makes it challenging to only learn emotion-related information from a multi-speaker database (Sisman and Li, 2018b; Zhou et al., 2020c). As shown in Fig. 5, phonetic posteriorgrams (PPGs) are utilized as the auxiliary input features for emotional voice conversion (Liu et al., 2020b). It is noted that PPGs are derived from an automatic speech recognition system, that are assumed to be speaker-independent and emotion-independent. With the phonetic information from PPGs, the proposed emotional voice conversion system demonstrates a better generalization ability with multi-speaker and multi-emotion speech data. In general, these frameworks have shown the effectiveness of multi-task learning or transfer learning, which represents a potential direction for future studies.

## 3. Databases in emotional voice conversion research

In this section, we present a survey of emotional speech databases available for development and evaluation for emotional voice conversion. We first give an overview of existing emotional speech databases in Table 1. Even though the given list of databases in Table 1 may not be exhaustive, it covers the most widely-used emotional speech databases to our best knowledge. There have been some surveys of existing emotional databases for speech emotion recognition that can be found in El Ayadi et al. (2011) and Akçay and Oğuz (2020). In this section, we would like to give a comprehensive overview of existing emotional speech databases from the standpoint of emotional voice conversion rather than emotion recognition. In order to motivate our *ESD* database, we provide an in-depth comparison of the state-of-the-art emotional speech databases in terms of lexical variability, language variability, speaker variability, confounding factors, and recording environment, as reported in Table 1. We also provide discussions and insights about the characteristics, strengths and weaknesses of these existing emotional speech databases in the following subsections.

### *3.1. Lexical variability*

It is known that the performance of speech synthesis and voice conversion systems is strongly dependent on the condition of the speech samples that are used during the training (Sisman et al., 2020). For both voice conversion and text-to-speech research, lexical content is especially important as it can affect the generalization ability of the speech synthesis system. In general, speech synthesis systems could benefit from the training data with a large lexical variety. However, in practice, such databases with a large variety of emotional utterances are not widely available, and most databases only contain limited utterances for each emotion.

We note that many of the existing emotional speech databases cover most of the emotion categories only with limited lexical variability. For example, the RAVDESS database (Livingstone and Russo, 2018) contains emotional speech data from 24 actors, and the actors are asked to read only 2 different sentences in a spoken and sung way. Each utterance is repeated with 2 different intensities (except for the neutral emotion). Similar to the RAVDESS database, the CREMA-D database (Cao et al., 2014) contains 12 different sentences recorded by 91 actors. The Berlin Emotional Speech Dataset (EmoDB) (Burkhardt et al., 2005) collects 10 sentences read by 10 speakers with 7 emotions. The Surrey Audio–Visual Expressed Emotion (SAVEE) (Jackson and Haq, 2014) database covers 14 speakers with 7 different emotions but only contains 480 utterances in total. JL-Corpus (James et al., 2018) is a collection of 4 New-Zealand English speakers uttering 15 sentences in 5 primary emotions with 2 repetitions, and another 10 sentences in 5 secondary emotions. In general, these databases are all of good quality and can be ideal to build a rule-based emotional voice conversion framework. However, they might not be suitable for building a comprehensive model for emotional speech, especially with current deep learning approaches, due to their limited variety of sentences.

There are also some emotional speech databases designed for lexical control. In the Toronto Emotional Speech Set (TESS) (Pichora-Fuller and Dupuis, 2020), 2 speakers speak 200 words with 7 different emotions in the carrier phrase ("Say the word …"). A recent VESUS database (Sager et al., 2019) is designed and released with over 250 distinct phrases, each read by ten actors in five emotional states. These databases mark valuable practice to understand the emotion variance in the word or phrase level, but may not be suitable to build a state-of-the-art emotional voice conversion framework that is usually data-driven. It is also known that emotion is hierarchical in nature and can be varying from different prosody levels, we believe it would be more ideal for emotional voice conversion research to study the emotion variance within the utterances (Zhou et al., 2020b; Xu, 2011; Latorre and Akamine, 2008). In general, due to the limited lexical variability, these databases may not be ideal to build an emotional voice conversion framework, especially a speaker-independent one, with state-of-the-art deep learning approaches. Therefore, it is needed to have an emotional speech database with a large lexical variability to boost the related studies, such as emotional voice conversion in a speaker-independent scenario.

### *3.2. Language variability*

Another limitation is the imbalanced language representation that can be observed from Table 1. We note that 9 databases out of 19 listed in Table 1 only contain English speech, while the others include German, French, Danish, Italian, Japanese and Chinese. Among them, 2 databases contain multilingual content of English and French, those are AmuS (El Haddad et al., 2017) and EmoV-DB (Adigwe et al., 2018). AmuS is a database which is designed to study the expression of amusement across English and French, and contains 3 h speech data in total. EmoV-DB contains emotional speech data performed by four English speakers and one Belgium French speaker in five emotion categories. Besides, there is an attempt to build multi-lingual emotional speech databases covering different languages including Chinese, English, Japanese, Korean and Russian (Wang et al., 2012). These

**Table 1**
An overview of existing emotional speech databases.

| Database | Language | Emotions | Size | Modalities | Content type | Remarks |
|---|---|---|---|---|---|---|
| RAVDESS (Livingstone and Russo, 2018) | EN | Neutral, happy, angry, sad, calm, fear, disgust | 24 actors (12 male and 12 female) x 2 sentences x 8 emotions x 2 intensities (normal, strong) | Audio/Visual | Acted | Limited lexical variability |
| CREMA-D (Cao et al., 2014) | EN | Neutral, happy, angry, sad, fear, disgust | 91 actors (48 male and 43 female) x 12 sentences x 6 emotions, 1 sentence is expressed with 3 intensities (low, medium, high), others are not specified | Audio/Visual | Acted | Limited lexical variability |
| EmoDB (Burkhardt et al., 2005) | GE | Neutral, happy, angry, disgust, bored | 10 speakers (5 male and 5 female) x 10 sentences x 7 emotions | Audio/Visual | Acted | Limited lexical variability |
| MSP-IMPROV (Busso et al., 2016) | EN | Neutral, happy, angry, sad | 12 actors, 9 h of data | Audio/Visual | Acted/Improvised | Contains external noise, and overlapping speech |
| IEMOCAP (Busso et al., 2008) | EN | Neutral, happy, angry, sad, frustrate | 10 speakers (5 male and 5 female), 12 h of data | Audio/Visual | Acted/Improvised | Contains external noise, and overlapping speech |
| DES (Engberg et al., 1997) | DA | Neutral, happy, angry, sad, surprise | 4 speakers (2 male and 2 female), 10 min of speech | Audio | Acted | Limited lexical variability |
| CHEAVD (Li et al., 2017) | CN | Neutral, happy, angry, sad, surprise, fear | 219 speakers, two hours emotional segments from films and TV shows | Audio/Visual | Improvised | Contains external noise |
| CASIA (Zhang et al., 2008) | CN | Neutral, happy, angry, sad, surprise, fear | 4 speakers (2 male and 2 female) x 500 utterances x 6 emotions | Audio/Visual | Acted | Limited speaker variability Not free to use |
| AmuS (El Haddad et al., 2017) | EN, FR | Neutral, amuse | 4 speakers (3 male and 1 female), about 3 h of data | Audio | Acted | Only single emotion |
| EmoV-DB (Adigwe et al., 2018) | EN, FR | Neutral, amuse, angry, sleepy, disgust | 4 English speakers (2 male and 2 female), 1 French speaker (male), about 5 h of data | Audio | Acted | Contains other non-verbal expressions (laughter, sigh) Incomplete public version Limited speaker variability |
| SAVEE (Jackson and Haq, 2014) | EN | Neutral, happy, sad, angry, surprise, disgust, fear | 4 speakers (all male), 480 utterances in total | Audio/Visual | Acted | Limited speaker variability (all male) Limited lexical variability |
| VESUS (Sager et al., 2019) | EN | Neutral, happy, angry, sad, fear | 10 actors (5 male and 5 female) x 250 distinct phrases | Audio | Acted | Phrases, not complete sentences |
| JL-Corpus (James et al., 2018) | EN | Neutral, happy, angry, sad, excite | 4 speakers (2 male and 2 female) x 5 emotions x 15 sentences x 2 repetitions; 4 speakers (2 male and 2 female) x 6 variants (questions/interrogative, apologetic, encouraging/reassuring, concerned, assertive, anxiety) x 10 sentences x 2 repetitions | Audio | Acted | Limited speaker variability Limited lexical variability |
| eNTERFACE'05 (Martin et al., 2006) | EN | Angry, disgust, fear, happy, sad, surprise | 42 speakers (34 male and 8 female) from 14 nationalities, 1116 video clips in total | Audio/Visual | Improvised | With different accents Limited lexical variability |
| TESS (Pichora-Fuller and Dupuis, 2020) | EN | Neutral, angry, sad, fear, disgust, surprise, pleased | 2 speakers (female) x 200 words x 7 emotions (in the carrier phrase "Say the word _") | Audio | Acted | Limited speaker variability Limited lexical variability |
| DEMoS (Parada-Cabaleiro et al., 2020) | IT | Neutral, angry, happy, sad, fear, disgust, surprise, guilty | 9365 emotional and 332 neutral samples produced by 68 native speakers (23 females, 45 males) | Audio | Improvised | Imbalanced setting for each emotion |
| EMOVO (Costantini et al., 2014) | IT | Neutral, disgust, fear, anger, joy surprise, sad | 6 actors (3 male and 3 female) x 14 sentences x 7 emotions | Audio | Acted | Limited speaker variability Limited lexical variability |
| VAM Corpus (Grimm et al., 2008) | GE | Valence, activation, dominance | 12 h of audio–visual recordings of a German TV talk show | Audio/Visual | Improvised | Contain other non-verbal expressions and external noise |
| JTES (Takeishi et al., 2016) | JP | Neutral, sad, joy, angry | 23 h and 31 min of 50 spoken sentences of 5 emotions acted by 100 speakers (50 male and 50 female) | Audio | Acted | Not publicly available |

Language abbreviations are used as follows: *EN* stands for English, *GE* stands for German, *DA* stands for Danish, *IT* stands for Italian, *JP* stands for Japanese, *FR* stands for French and *CN* stands for Chinese (Mandarin).

databases mark the valuable practice that allow for the study of emotion expression across different languages within the same database. There are also some emotional speech databases designed for those languages which are under-explored, such as Italian (Costantini et al., 2014), Finnish (Seppänen et al., 2003), Basque (Saratxaga et al., 2006) and Polish (Staroniewicz and Majewski, 2009). However, the imbalance of the emotional speech data between English and other languages still makes it tough to learn a cross-lingual emotion representation for emotional voice conversion.

In general, the omnipresence of English in existing databases is mainly due to historical reasons, and many other languages are considered as low-resource (Tits et al., 2019a). Since the lexical content is also constrained by the language, there have been some efforts to build a cross-lingual (Sisman et al., 2019a) or multi-lingual (Nekvinda and Dušek, 2020) speech synthesis system in current literature. However, the lack of multi-lingual emotional speech databases in the literature makes the related studies in emotional speech synthesis or emotional voice conversion more difficult, thus motivates us to build a multi-lingual emotional speech database that can achieve a balance between English and other languages, and suitable for a study of emotional expression across different languages.

### 3.3. Speaker variability

Emotional expression presents both speaker-dependent and speaker-independent characteristics at the same time (Arnold, 1960). Previous studies on emotion recognition (Kotti and獎Paternò, 2012) and emotional voice conversion (Zhou et al., 2020c) have shown a better generalization ability by learning a speaker-independent emotion representation over a large multi-speaker emotional corpus. These studies indicate the importance of a database with a large speaker variety for emotional speech synthesis (Choi et al., 2019) and emotional voice conversion (Zhou et al., 2020c).

We note that many existing emotional speech databases lack speaker variability, thus not optimal for the study of speaker-independent emotional voice conversion. For example, DES (Engberg et al., 1997) contains emotional speech data from four Danish speakers. Similar to DES, JL-Corpus (James et al., 2018) consists of emotional speech data from four New Zealand English speakers, and the EmoV-DB database (Adigwe et al., 2018) is built with about 5 h of multi-lingual speech data covering 5 different emotions and recorded by 4 English speakers and a French speaker. SAVEE (Jackson and Haq, 2014) has emotional utterances spoken by 14 speakers, but all of them are male. These databases contain high-quality emotional speech data that can be used to build an emotional voice conversion for a single speaker, that namely speaker-dependent one. However, they may not be optimal when building a speaker-independent emotional voice conversion framework, due to their limited speaker variability.

### 3.4. Confounding factor

It is known that emotion can be varying from individuals and their backgrounds (Kotti and Paternò, 2012; Fersini et al., 2009; Arnold, 1960; Dai et al., 2009). As a para-linguistic factor, emotion is complex with multiple speech characteristics, such as age and accent. In emotional voice conversion, we would like to convert the emotional state of the individual, and carry other para-linguistic characteristics from the source to the target. In the study of speaker-independent emotional voice conversion, we always expect an emotional speech database with a large variety of speakers, and other confounding factors should be controlled at the same time.

We note that there are some emotional speech databases with a large variety of speakers but not suitable to build a speaker-independent emotional voice conversion framework, due to their confounding factors. For example, the eNTERFACE'05 (Martin et al., 2006) contains English speech data of 42 different speakers from 14 nationalities. However, the underlying accents in the speech are unwanted for emotional voice conversion. IEMOCAP (Busso et al., 2008) is collected from conversations, and contains confounding factors such as laughter and sigh. These confounding factors can affect the conversion performance, thus they are unwanted in designing an emotional voice conversion database.

### 3.5. Recording environment

Unlike the robustness of emotion recognition systems could benefit from the external noises, the performance of speech synthesis is hugely dependent on the quality of the audio samples. The speech samples with low signal-to-noise ratio (SNR) or low sampling rate could harm the quality of the synthesized audios. Therefore, we always require a typical indoor environment with professional recording devices during the audio collection for speech synthesis tasks. We note that there are some emotional speech databases with a large amount of speech data from a large variety of speakers. The speech data is usually collected from dynamic conversations, films or TV shows, and evaluated by subjects. However, these databases might not be optimal to build speech synthesis systems because of their recording environments.

As listed in Table 1, there are some databases containing a large number of diverse sentences of emotional data. For example, IEMOCAP (Busso et al., 2008) contains audio–visual recordings of 5 sessions, each of which is displayed by a pair of speakers (male and female) in scripted and improvised scenarios. In total, the database contains 10 speakers and 12 h of data. Similar to IEMOCAP, MSP-IMPROV (Busso et al., 2016) consists of 6 sessions and 9 h of audio–visual data from 12 actors. These databases are both evaluated in terms of emotion category (Ekman, 1992) and emotional dimensions (Posner et al., 2005). CHEAVD (Li et al., 2017) is a large-scale emotional speech database for Mandarin Chinese that contains 219 speakers. In CHEAVD, all the audio–visual data are segmented from films and TV shows. We note that these databases have been widely used for speech emotion recognition because they have a large variability of speakers and lexical content. However, they are not suitable for synthesis purpose due to the overlapping speech and some external noise. Their recording setups could affect the performance of synthesized audios, thus these databases are not ideal choices for emotional speech synthesis and emotional voice conversion.

## 4. ESD database

The *ESD* database was recorded with the aim to provide the community with a large emotional speech database with a sufficient variety of speaker and lexical coverage. In this section, we first explain how the *ESD* database overcomes the weakness of existing emotional speech databases. We then present the design details and provide a comprehensive analysis of the *ESD* database. We also evaluate the *ESD* database with a speech emotion recognition framework, and report its classification performance, and visualize the emotion embeddings that are learnt from the network.

### 4.1. Addressing the research need

According to the content type, emotional speech databases can be divided into two types: acted and improvised, as shown in Table 1. We note that both types of databases have their pros and cons (Juslin and Laukka, 2001). For example, the actors may create stereotypical portrayals of emotions in acted emotional speech (Bachorowski and Owren, 1995; Kappas and Hess, 2011). As for improvised speech databases, emotions are induced in the speaker under naturalistic conditions. However, it is hard to induce strong and well-differentiated emotions in this manner (Johnstone and Scherer, 2000; Banse and Scherer, 1996). We note that acted speech databases are widely used in emotional speech synthesis (Livingstone and Russo, 2018; Burkhardt et al., 2005; Adigwe et al., 2018), and improvised speech databases are frequently used in speech emotion recognition (Busso et al., 2008; Li et al., 2017). To be consistent with the literature on emotional speech synthesis, we would like to design a database with parallel and acted emotional speech. During the database design, we require the speakers to act with different emotions with the same linguistic content.

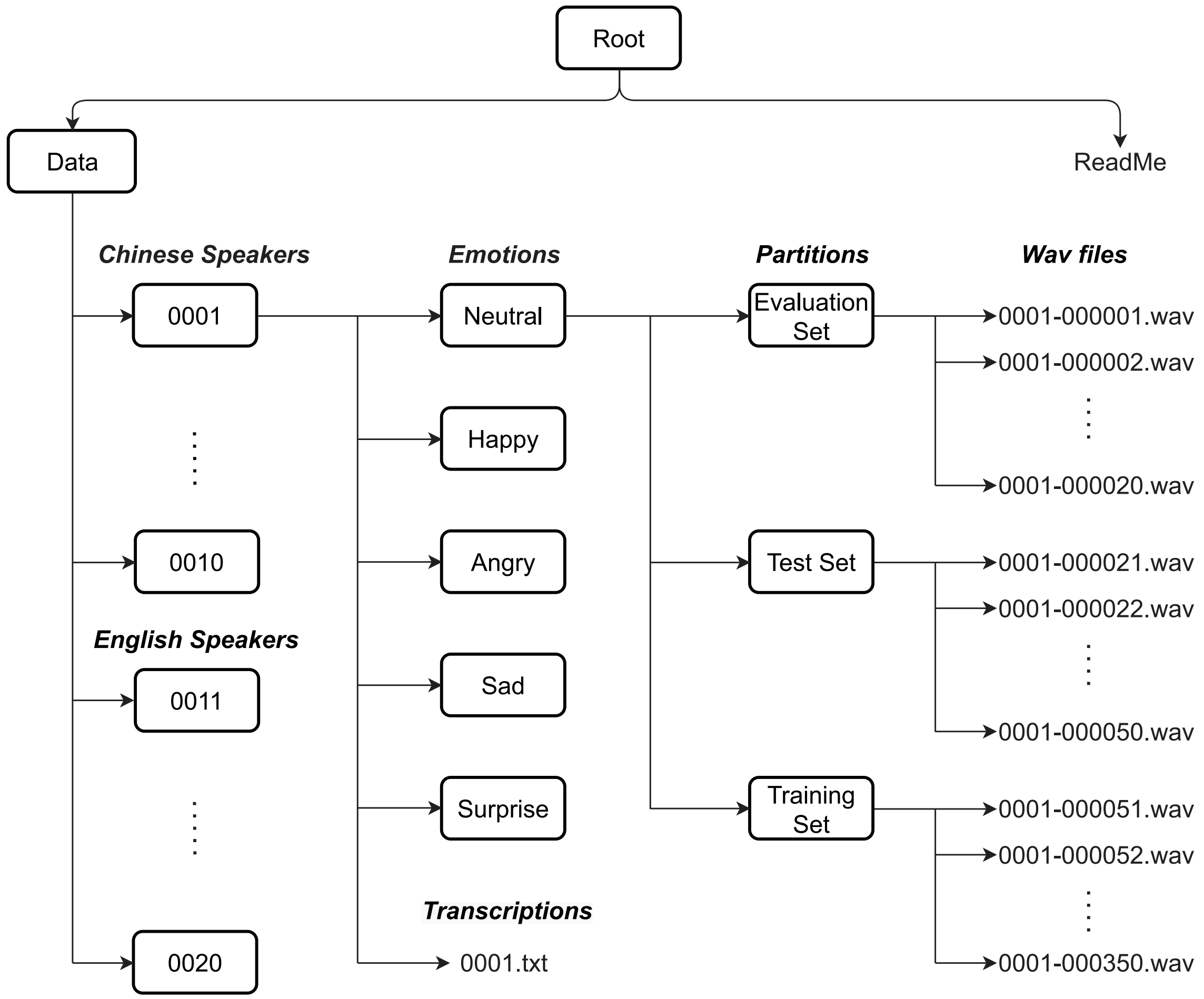

**Fig. 6.** Schematic representation of directory organization of the *ESD* database.

The *ESD* database is designed to fill the gaps left by the existing emotional speech databases as discussed in Section 3, in terms of *lexical variability*, *language variability*, *speaker variability*, *confounding factors*, and *recording environment*. Next, we will explain how *ESD* database addresses the above five aspects.

1. Lexical variability
   As mentioned in Section 3, most the existing emotional speech databases do not have sufficient lexical coverage. We also note that most of voice conversion databases, such as VCC 2016 (Toda et al., 2016), usually contain hundreds of utterances for each speaker for lexical variability. With parallel utterances, it also makes possible to compare the same sentence given by the speakers expressing different emotions. Therefore, we propose to record 350 parallel utterances for each emotion from each speaker in the *ESD* database.
2. Language variability
   The *ESD* database includes Chinese and English speech data at the same time, which is therefore naturally multi-lingual. The balanced setting of both languages allows us to study emotional expression across different languages and cultures. The *ESD* database also can be easily used to build a cross-lingual emotional voice conversion or speaker voice conversion framework.
3. Speaker variability
   We propose to have emotional speech data containing 10 Chinese speakers and 10 English speakers with a balanced gender setting, that has a larger number of speakers than previous databases (El Haddad et al., 2017; Adigwe et al., 2018; James et al., 2018). Furthermore, the speakers are instructed to speak hundreds of parallel utterances in different emotions. In this way, we expect the *ESD* database to facilitate the speaker-independent emotional voice conversion studies.
4. Confounding factor
   To control the confounding factors in speech, such as age, accents as discussed in Section 3, we propose to have all speakers in the same age group of 25–35. The speakers are instructed to refrain from any confounding expressions, such as laughter or sigh. All English speakers speak North American English, while all Chinese speakers speak standard Mandarin.
5. Recording environment
   To ensure an ideal recording environment, all the speech data in *ESD* database is recorded in a typical indoor environment with an SNR of above 20 dB and a sampling frequency of 16 kHz. It ensures that the *ESD* database are suitable to build a state-of-the-art emotional voice conversion framework.

### *4.2. Design*

The *ESD* database is recorded by 10 native English speakers and 10 native Chinese speakers. Each speaker contributes 350 utterances,

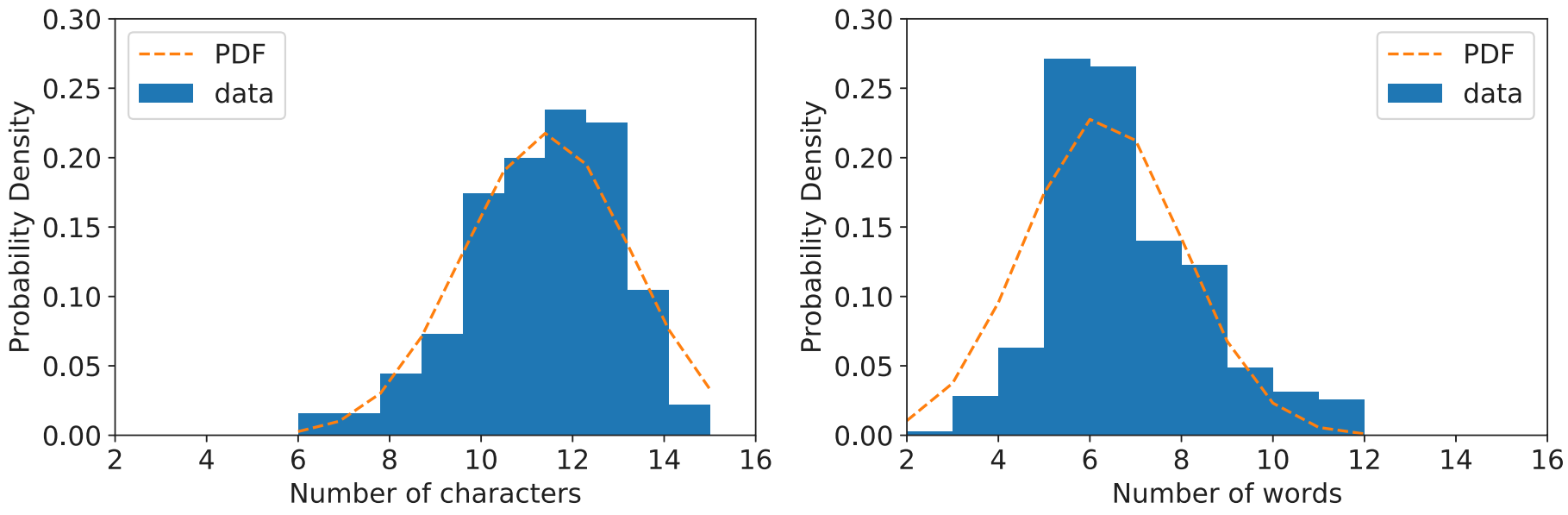

(a) Chinese (b) English

**Fig. 7.** The histograms of probability density of the number of characters/words per utterance for Chinese and English speech data in *ESD* database. The statistics of characters are reported for Chinese, and the statistics of words are reported for English. Probability density function (PDF) is estimated with Gaussian density function.

**Table 2**
There are 350 English or Chinese utterances for each speaker, and in each emotion. We partition each speaker-emotion set into an evaluation set, a test set, and a training set.

| Set | Number of utterances | |
|---|---|---|
| | English | Chinese |
| Evaluation set | 20 | 20 |
| Test set | 30 | 30 |
| Training set | 300 | 300 |

covering 5 emotion categories, namely *Neutral*, *Happy*, *Angry*, *Sad*, and *Surprise*. It is made available for research purpose upon request.[2]

#### 4.2.1. Partition

We propose to partition the *ESD* database as shown in Table 2. We partition the 350 parallel utterances of a speaker by emotion into three non-overlapping sets, which are evaluation, test, and training set. We use the first 20 utterances for the evaluation set, the following 30 utterances for the test set, and the rest of the utterances for the training set.

#### 4.2.2. Organization

The directory organization of *ESD* database is shown in Fig. 6. The *Data* folder and *ReadMe* are under the *Root* directory. In the Data folder, we have 20 sub-folders for speakers including 10 Chinese speakers (0001–0010) and 10 English speakers (0011–0020). In each speaker's folder, we have a text file for transcriptions, and 5 sub-folders corresponding to five emotion categories. For each emotion folder, we partition all the speech data into three sub-folders, that are *Evaluation Set*, *Test Set*, and *Training Set*.

### 4.3. Statistics

A summary of the statistics of the *ESD* database is presented in Table 3. In order to give a comprehensive analysis of *ESD* database, the statistics of lexical variability, duration, and F0 are further presented and discussed in the following.

#### 4.3.1. Lexicon

We report the statistics of English words for English speech data, and Chinese characters for Chinese speech data respectively. For each language, 350 parallel utterances are collected in five different emotions, resulting in a total of 1750 utterances for each speaker. We collect speech from 10 speakers for both Chinese and English.

For Chinese, the 1750 utterances have a total character count of 20,025 Chinese characters and 939 unique Chinese characters. They are commonly used in daily communication. On average, there are 11.5 characters in each Chinese utterance, as shown in Fig. 7(a).

For English, the 1750 utterances have a total word count of 11,015 words and 997 unique lexical words. They are the commonly used English words in daily communication. On average, there are about 6.31 words in each English utterance, as shown in Fig. 7(b).

In our work, 350 utterances are chosen to make sure that the text scripts have a large lexical coverage while reducing the repetition of word or character. The lexicon size in the ESD database is close to that for a person to speak and understand a language (Sagar-Fenton and McNeill, 2018).

#### 4.3.2. Duration

Speech duration is the manifestation of speaking rhythm. In Fig. 8, the probability density histograms of the utterance duration of *ESD* database are presented by emotion and language. For both English and Chinese, we observe that the utterance duration of those emotions with lower arousal and negative valence values such as sad tends to have a higher mean and a higher variance than that of emotions with higher arousal and positive valence such as happy (excited). This suggests that there could be some common codes of emotional expression across different languages, which motivates a multi-lingual study on emotional voice conversion.

It is reported that the average utterance and word duration is 2.76 s and 0.44 s respectively for English speech data, and 3.22 s and 0.28 s respectively for Chinese. We notice that Chinese speech has a longer average utterance duration but shorter character duration than English speech, that is mainly because the number of characters in a Chinese sentence is usually larger than the number of words in an English one, as observed in Fig. 7. We also note that the average utterance duration of *ESD* database is similar to that of most publicly available voice conversion databases in the literature (Toda et al., 2016; Lorenzo-Trueba et al., 2018b; Zhao et al., 2020).

#### 4.3.3. Fundamental frequency (F0)

Fundamental frequency (F0) is an essential prosodic descriptor of speech. In Table 4, we summarize the mean and standard variance of F0 over all the utterances by emotion and by gender. The F0 values are extracted by WORLD vocoder (Morise et al., 2016).

As shown in Table 4, we observe that the female speakers typically have a higher F0 value and standard variance than the male speakers within each emotion category. For both male and female speakers, happy and surprise tend to have a higher mean and variance of F0 than the others. The F0 of the angry category is generally higher than neutral and sad, but it has a lower variance than happy and surprise. The F0 of the sad and neutral categories have lower values, while sad has a slightly larger F0 variance than neutral. These observations generally conform with the findings in the prior studies (Xue et al., 2018).

[2] **ESD database**: https://github.com/HLTSingapore/Emotional-Speech-Data.

**Table 3**
A summary of the statistics of *ESD* database.

| Parameter | Chinese | | | | | | English | | | | | |
|---|---|---|---|---|---|---|---|---|---|---|---|---|
| | Neu | Ang | Sad | Hap | Sur | All | Neu | Ang | Sad | Hap | Sur | All |
| # speakers | 10 | 10 | 10 | 10 | 10 | 10 | 10 | 10 | 10 | 10 | 10 | 10 |
| # utterances per speaker | 350 | 350 | 350 | 350 | 350 | 1750 | 350 | 350 | 350 | 350 | 350 | 1750 |
| # unique utterances | 350 | 350 | 350 | 350 | 350 | 350 | 350 | 350 | 350 | 350 | 350 | 350 |
| # characters/words per speaker | 4005 | 4005 | 4005 | 4005 | 4005 | 20,025 | 2203 | 2203 | 2203 | 2203 | 2203 | 11,015 |
| # unique characters/words | 939 | 939 | 939 | 939 | 939 | 939 | 997 | 997 | 997 | 997 | 997 | 997 |
| Avg. utterance duration [s] | 3.23 | 2.68 | 4.04 | 2.84 | 3.32 | 3.22 | 2.61 | 2.80 | 2.98 | 2.70 | 2.73 | 2.76 |
| Avg. character/word duration [s] | 0.28 | 0.23 | 0.35 | 0.25 | 0.29 | 0.28 | 0.41 | 0.44 | 0.47 | 0.43 | 0.43 | 0.44 |
| Total duration [s] | 11,305 | 9380 | 14,140 | 9940 | 11,620 | 56,385 | 9135 | 9800 | 10,430 | 9450 | 9555 | 48,370 |

Emotion abbreviations are used as follows: *Neu* stands for neutral, *Ang* stands for angry, *Sad* stands for sad, *Hap* stands for happy and *Sur* stands for surprise. The statistics of characters are reported for Chinese, and the statistics of words are reported for English.

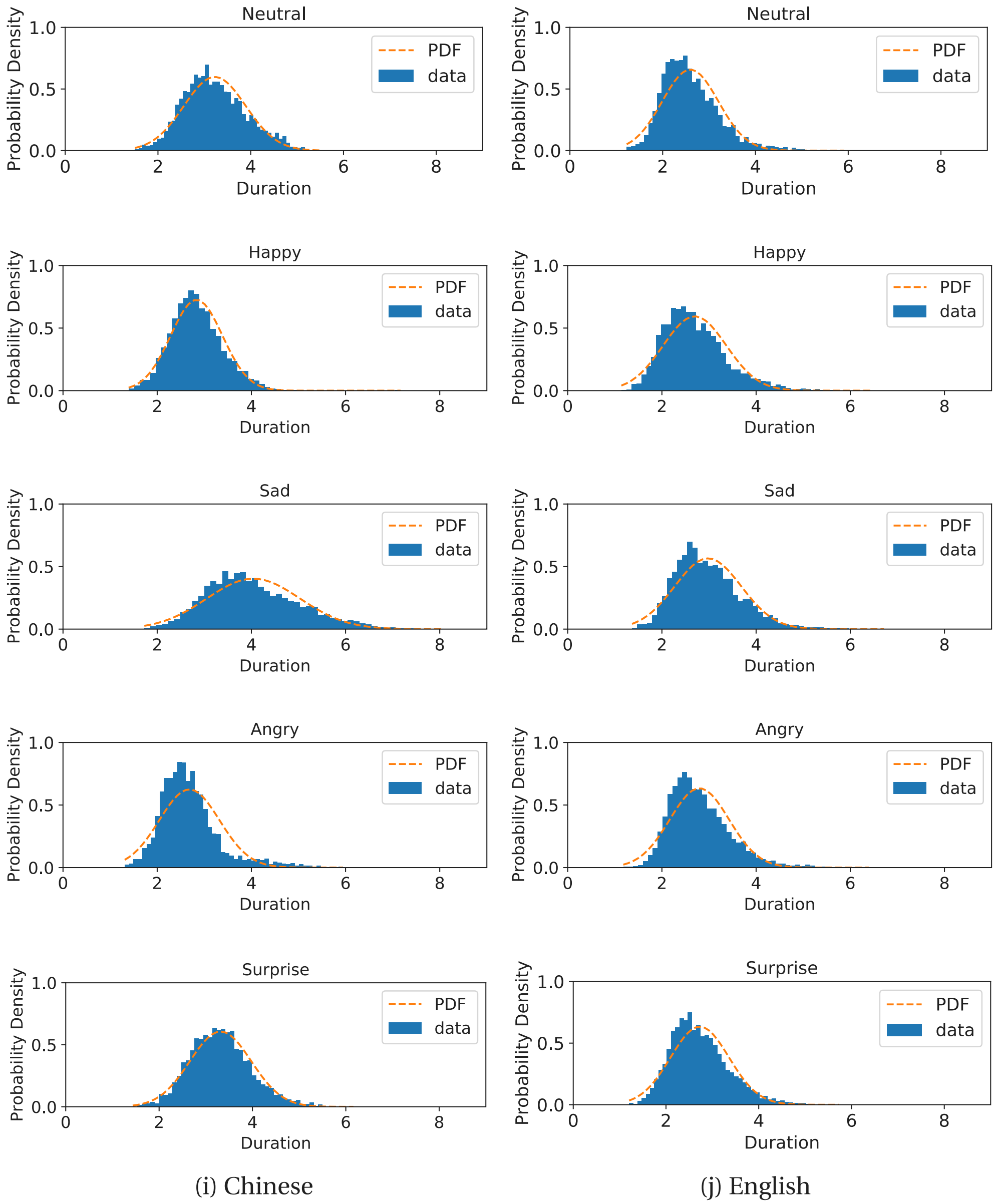

**Fig. 8.** Probability density histograms of the utterance duration [s] of five emotions for Chinese and English speech data. The histograms are reported with fifty bins. Probability density function (PDF) is estimated with Gaussian density function.

### *4.4. Emotion classification*

To validate the quality of emotional expression in *ESD* database, we develop a state-of-the-art speaker-independent speech emotion recognition system. We follow the same data partition presented in Fig. 6. We use the training data from all the speakers to develop a speech emotion recognition system for Chinese and English respectively. We first extract frame-level features with the sliding window of 25 ms and a shift of 10 ms. For each frame, we then extract 312-dimensional feature vectors with openSMILE (Eyben et al., 2010), including zero-crossing

**Table 4**
A summary of mean and standard variance (std) of F0.

| Parameter | Neutral | | Angry | | Sad | | Happy | | Surprise | |
|---|---|---|---|---|---|---|---|---|---|---|
| | Male | Female | Male | Female | Male | Female | Male | Female | Male | Female |
| ***(English speech data)*** | | | | | | | | | | |
| F0 mean [Hz] | 107.60 | 148.92 | 134.88 | 167.28 | 114.64 | 154.39 | 142.86 | 206.98 | 154.21 | 211.52 |
| F0 std [Hz] | 26.02 | 37.64 | 45.38 | 54.08 | 37.46 | 31.69 | 51.77 | 68.69 | 61.26 | 82.19 |
| ***(Chinese speech data)*** | | | | | | | | | | |
| F0 mean [Hz] | 91.22 | 156.59 | 144.34 | 193.23 | 102.18 | 199.79 | 142.27 | 221.45 | 156.09 | 226.63 |
| F0 std [Hz] | 19.88 | 29.63 | 41.26 | 53.57 | 21.67 | 40.98 | 46.99 | 64.47 | 50.52 | 71.07 |

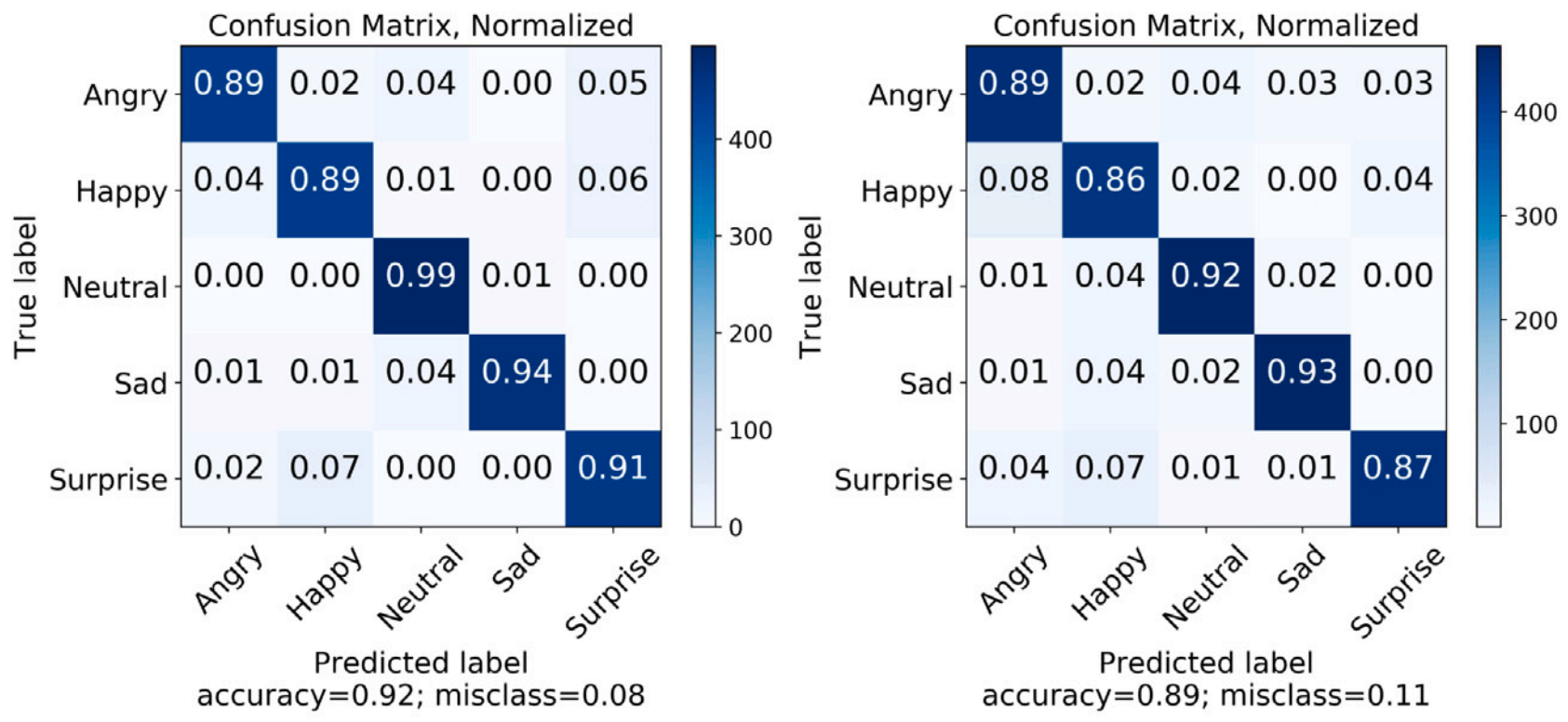

(a) Chinese (b) English

**Fig. 9.** Confusion matrix of Chinese and English test data in a SER experiment on the *ESD* database. The diagonal entries represent the recall rates of each emotion.

**Table 5**
A comparison of the MCD [dB] values of Zero effort, CycleGAN-EVC (Zhou et al., 2020b), and VAWGAN-EVC (Zhou et al., 2021b) for four emotion conversion pairs. The lower value of MCD indicates the better conversion performance.

| MCD [dB] | Neutral-to-Angry | Neutral-to-Happy | Neutral-to-Sad | Neutral-to-Surprise |
|---|---|---|---|---|
| Zero effort | 6.47 | 6.64 | 6.22 | 6.49 |
| CycleGAN-EVC | 4.57 | 4.46 | 4.32 | 4.68 |
| VAWGAN-EVC | 5.89 | 5.73 | 5.31 | 5.67 |

rate, voicing probability, mel-frequency cepstral coefficients (MFCCs) and mel-spectrum.

The SER model consists of a LSTM layer, followed by a ReLU-activated fully connected (FC) layer with 256 nodes. Dropout is applied on the LSTM layer with a keep probability of 0.5. Finally, the resulting 256-features vector is fed to a softmax classifier, which is a FC layer with 5 outputs. We follow the data partition in Table 2 in this experiment, and present a confusion matrix to report the SER results. As shown in Fig. 9, we observe an overall SER accuracy of 92.0% and 89.0% respectively for Chinese and English systems. The results suggest that the speech in the *ESD* database is rendered with high level of emotion consistency.

Deep emotional features are intermediate activations of the SER system, that characterize emotion states (Liu et al., 2020c; Zhou et al., 2021c). We randomly select 20 utterances for each emotion from the test set and visualize their deep emotional features using the t-distributed stochastic neighbor embedding (t-SNE) algorithm (Maaten and Hinton, 2008) in a two-dimensional plane, as shown in Fig. 10. The results suggest that the deep emotional features derived from *ESD* serve as an excellent descriptor of speech emotion states. Encouraged by this, we will study the use of deep emotional features as the emotion code for emotional voice conversion.

## 5. Emotional voice conversion on ESD

To provide a reference benchmark, we conduct emotional voice conversion experiments on the *ESD* database using state-of-the-art techniques.

### 5.1. Experimental setup

We implement two state-of-the-art emotional voice conversion frameworks, that are (1) *CycleGAN-EVC* (Zhou et al., 2020b) (see Fig. 3), and (2) *VAWGAN-EVC* (Zhou et al., 2021b) (see Fig. 4). The former learns the pair-wise emotion domain translation through the CycleGAN network itself, while the latter relies on the emotion codes to control the conversion. We made available the implementation of the two frameworks for ease of reference.[3,4] It is noted that ESD also can be easily used for speaker-independent EVC.

We choose the emotional speech data of one male speaker (‘*0013*’) from *ESD* database to carry out all the experiments. We conduct emotion conversion from neutral to angry, happy, sad and surprise respectively. For each emotion, there are 350 parallel utterances in total. In all experiments, we follow the same partition of evaluation, test, and training set described in Section Section 4.2.1. For each baseline framework, we train one model for each emotion pair, i.e, neutral to angry (‘*Neu-Ang*’), neutral to happy (‘*Neu-Hap*’), neutral to sad (‘*Neu-Sad*’), and neutral to surprise (‘*Neu-Sur*’), with the same set of training data. It is noted that there is no frame-alignment procedure in all the experiments, thus both frameworks do not require parallel training data.

3 **CycleGAN-EVC**: https://github.com/KunZhou9646/emotional-voice-conversion-with-CycleGAN-and-CWT-for-Spectrum-and-F0.

4 **VAWGAN-EVC**: https://github.com/KunZhou9646/Speaker-independent-emotional-voice-conversion-based-on-conditional-VAW-GAN-and-CWT.

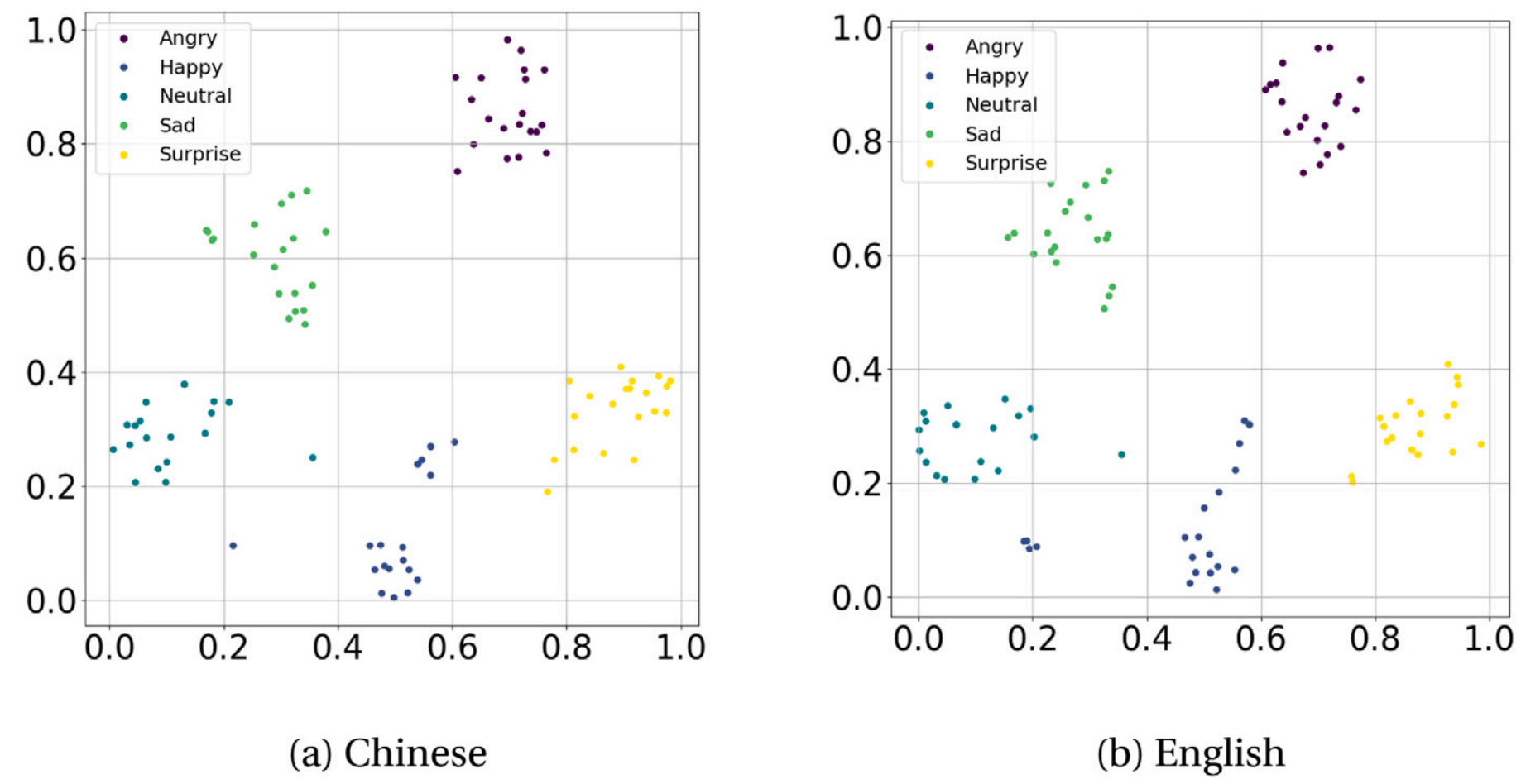

**Fig. 10.** Visualizations of the deep emotional features for 20 utterances.

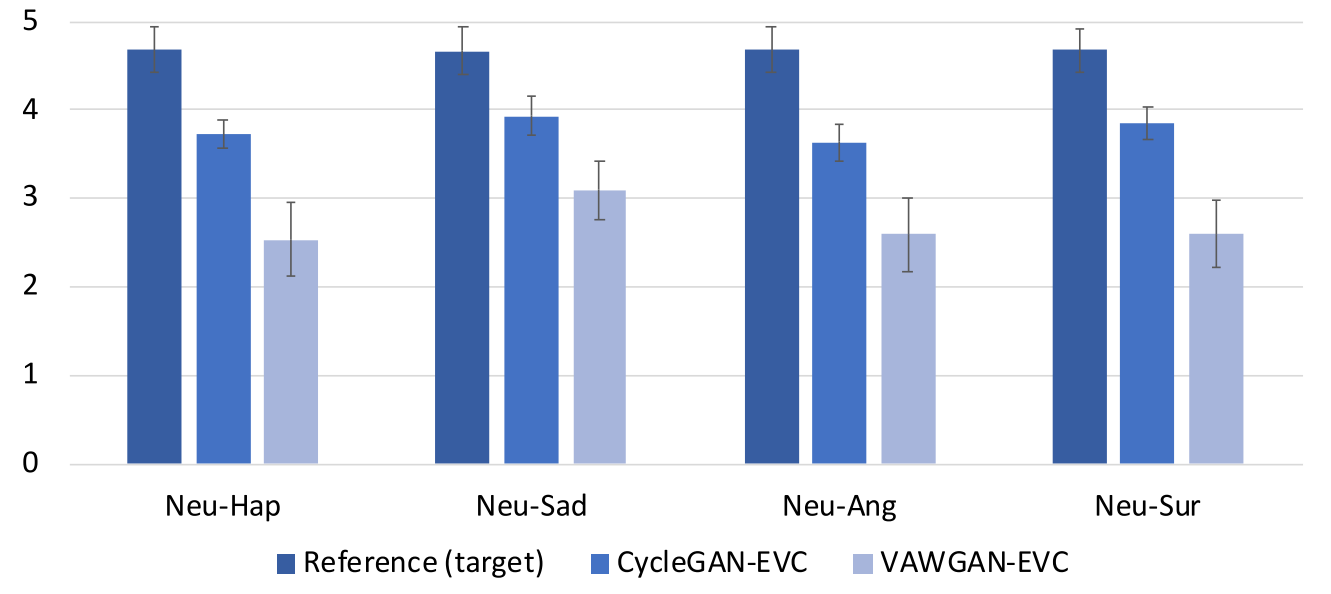

**Fig. 11.** MOS test results for speech quality with 95% confidence interval. This test aims to compare the speech quality of reference target audio and the converted audio with CycleGAN-EVC and VAWGAN-EVC.

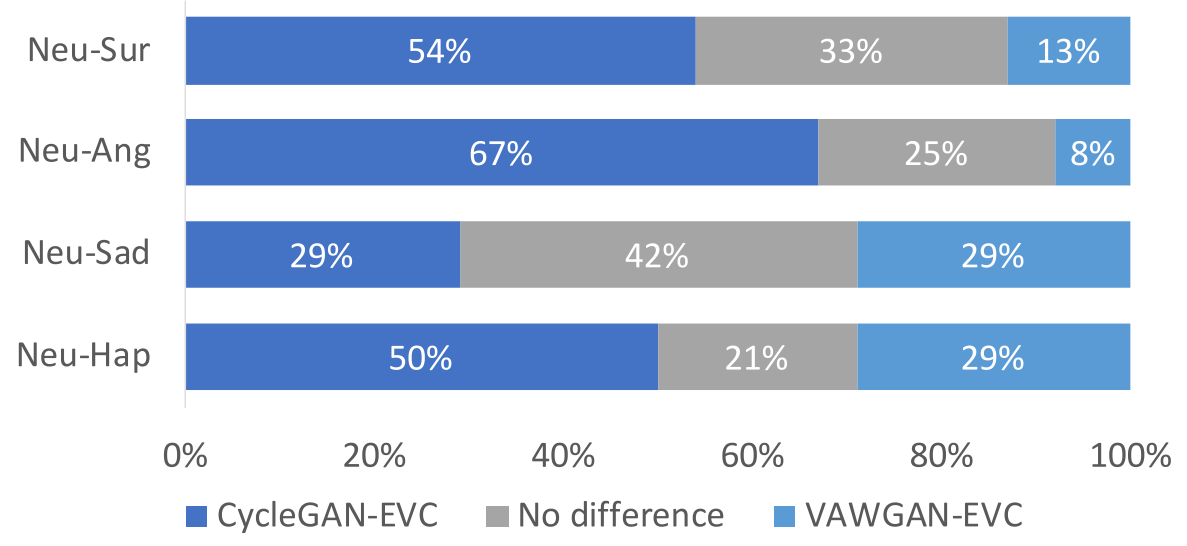

**Fig. 12.** XAB preference test results for emotion similarity. This test is conducted to compare the performance of CycleGAN-EVC and VAWGAN-EVC in terms of the emotion similarity with the reference target audio.

### 5.2. Objective evaluation

For both frameworks, we derive 24-dimensional Mel-cepstral coefficients (MCEPs) from the spectrum, calculate their Mel-cepstral distortion (MCD) values (Kubichek, 1993) between the converted and target MCEPs, and report in Table 5. In Table 5, *Zero effort* denotes the case where we directly compare the source with the target without any conversion. We report *Zero effort* as a contrastive reference. As shown in Table 5, we observe that both CycleGAN-EVC and VAWGAN-EVC achieve lower MCD values than *Zero effort*, which is encouraging. Moreover, the CycleGAN-EVC consistently outperforms VAWGAN-EVC for all conversion pairs.

### 5.3. Subjective evaluation

We further conduct listening experiments to assess the performance of CyleGAN-EVC and VAWGAN-EVC in terms of speech quality and emotion similarity. 11 subjects participate in all the listening experiments, and each of them listens to 72 utterances in total.

We first report the mean opinion score (MOS) of reference target speech, CycleGAN-EVC and VAWGAN-EVC to assess the overall speech quality, which covers two aspects: (1) How the linguistic information and speaker identity is preserved, and (2) the naturalness of the converted speech. During the listening test, the listeners are asked to rate the quality of speech audio with a 5-point scale: 5 for excellent, 4 for good, 3 for fair, 2 for poor, and 1 for bad. As shown in Fig. 11, CycleGAN-EVC outperforms VAWGAN-EVC for all the emotion conversion pairs.

We also conduct an XAB preference test to assess the emotion similarity, where the subjects are asked to choose the speech sample which sounds closer to the reference in terms of emotional expression. As shown in Fig. 12, we observe that CycleGAN-EVC outperforms VAWGAN-EVC for Neu-Sur, Neu-Ang and Neu-Hap, and achieves comparable results with VAWGAN-EVC for Neu-Sad.

### 5.4. Discussion

We have evaluated two recent proposed EVC frameworks on the *ESD* database. We observe that CycleGAN-EVC achieves better performance than VAWGAN-EVC with a limited amount of data, especially in terms of speech quality. However, with the same amount of training data, the training of CycleGAN-EVC takes nearly 72 h on one GeForce GTX 1080 Ti, while that of VAWGAN-EVC only takes about 7 h with the same experimental setup. On the other hand, VAWGAN-EVC (Zhou et al., 2020c) allows for the conversion of many-to-many emotions, which is more controllable than CycleGAN-EVC, that can be extended for speaker-independent emotional voice conversion.

To evaluate the performance of the emotional expression in the converted speech, we regard the target speech emotion as the ground truth in emotional similarity experiment. We note that there exists a mismatch between the listener assessment and the speaker's intention for emotion, which will be studied in our future work.

Through the EVC experiments, we seek to evaluate the usability of the *ESD* database for emotional voice conversion studies. To our delight, both objective and subjective evaluations confirm that the *ESD* database provides an excellent shared task of emotional voice conversion under a controlled environment. The speech samples are publicly available.[5]

[5] **Speech Samples**: https://hltsingapore.github.io/ESD/demo.html.

## 6. Other voice conversion and text-to-speech on ESD

We have studied the use of the *ESD* database for emotional voice conversion. We also would like to highlight that the database can be used in other voice conversion and text-to-speech studies as well.

### 6.1. Speaker vs. emotional voice conversion

Typically voice conversion is used to describe speaker voice conversion, which seeks to convert speaker identity from one to another without changing the linguistic content (Mohammadi and Kain, 2017; Sisman et al., 2020), as shown in Fig. 1(b). In general, speaker voice conversion includes mono-lingual and cross-lingual voice conversion (Sisman et al., 2020). Cross-lingual voice conversion is a more challenging task because the source and target speakers have different native languages (Abe et al., 1990, 1991; Erro and Moreno, 2007; Du et al., 2020). In VCC 2020, cross-lingual voice conversion becomes one of the main tasks (Zhao et al., 2020). It is noted that a multi-lingual speech database is necessary for cross-lingual voice conversion.

We note that speaker voice conversion and emotional voice conversion are also different in many ways. For example, the former considers prosody-related features as speaker-independent and mostly focuses on spectrum conversion. However, emotion is the interplay of acoustic attributes concerning spectrum and prosody (Arnold, 1960). Therefore, in the latter, prosody conversion requires the same level of attention as spectral conversion. The *ESD* database offers an invaluable addition to extend the scope of the current voice conversion study.

### 6.2. Voice conversion databases vs. ESD

Unlike emotional voice conversion, there are abundant speech databases for speaker voice conversion. For example, VCC 2016 database (Toda et al., 2016) contains parallel speech data from 5 male speakers and 5 female speakers, with each speaker speaking 216 utterances in total. VCC 2018 database (Lorenzo-Trueba et al., 2018b) contains both parallel and non-parallel utterances from 6 male speakers and 6 females speakers. VCC 2020 database (Zhao et al., 2020) is a multi-lingual database that contains 4 different languages speech data. Other databases, such as the CMU-Arctic database (Kominek and Black, 2004) and the VCTK databases (Veaux et al., 2016) are also popular for voice conversion research. There are other large speech databases as well, such as LibriTTS (Zen et al., 2019) and VoxCeleb (Nagrani et al., 2020). LibriTTS corpus has 585 h of transcribed speech data uttered by a total of 2456 speakers, while the VoxCeleb database is further larger and consists of about 2800 h of untranscribed speech from over 6000 speakers. These two databases are mostly used to train a noise-robust speaker encoder to improve the speaker voice conversion performance.

We note that the speech in the above databases is emotion-free, which is not suitable for emotional voice conversion. The multi-speaker and cross-lingual nature of *ESD* database make possible new studies beyond the existing speaker voice conversion databases.

### 6.3. Emotional text-to-speech on ESD

For text-to-speech studies, there are many large-scale and high-quality speech databases, such as the LJ-Speech database (Ito and Johnson, 2017) and the Blizzard Challenge databases (Zhou et al., 2020a; Wu et al., 2019; King and Karaiskos, 2013). However, the emotional text-to-speech studies have been suffering from the lack of suitable emotional speech database as discussed in Section 3. Due to the lack of data, researchers mostly use internal databases for emotional text-to-speech studies (Um et al., 2020; Lei et al., 2021; Kwon et al., 2019; Zhu et al., 2019). We note that these internal databases are not publicly available and limit the related research in emotional text-to-speech.

The *ESD* database is designed to fill the gap and can be easily used for various text-to-speech tasks, especially for emotional text-to-speech. To our delight, there are already a few successful attempts (Zhou et al., 2021a; Liu et al., 2021) to build emotional text-to-speech frameworks on the *ESD* database. In Zhou et al. (2021a), researchers propose a joint training framework of emotional voice conversion and emotional text-to-speech. In Liu et al. (2021), an emotional text-to-speech framework with reinforcement learning is built and evaluated through the comparative study with several state-of-the-art emotional text-to-speech frameworks (Li et al., 2021; Cai et al., 2021) on the *ESD* database. These studies have all shown the effectiveness, which further show the promise of the *ESD* database on the emotional text-to-speech task.

## 7. Conclusion

This paper provides a comprehensive overview of the recent research and existing emotional speech databases for emotional voice conversion. To our best knowledge, this paper is the first overview paper that covers emotional voice conversion research and databases in recent years. Moreover, we release the *ESD* database and make it publicly available. The *ESD* database represents one of the largest emotional speech databases in the literature. By reporting several experiments on the *ESD* database, this paper provides a reference benchmark for emotional voice conversion studies that represent the state-of-the-art.

## CRediT authorship contribution statement

**Kun Zhou:** Conception and design of study, Acquisition of data, Analysis and/or interpretation of data, Drafting the manuscript, Revising the manuscript critically for important intellectual content. **Berrak Sisman:** Conception and design of study, Acquisition of data, Revising the manuscript critically for important intellectual content. **Rui Liu:** Drafting the manuscript, Revising the manuscript critically for important intellectual content. **Haizhou Li:** Revising the manuscript critically for important intellectual content.

## Declaration of competing interest

The authors declare that they have no known competing financial interests or personal relationships that could have appeared to influence the work reported in this paper.

## Acknowledgments

The research by Berrak Sisman and Rui Liu is funded by SUTD Start-up Grant Artificial Intelligence for Human Voice Conversion (SRG ISTD 2020 158) and SUTD AI Grant - Thrust 2 Discovery by AI (SG-PAIRS1821).

The research by Kun Zhou and Haizhou Li is supported by the Science and Engineering Research Council, Agency of Science, Technology and Research (A*STAR), Singapore, through the National Robotics Program under Human-Robot Interaction Phase 1 (Grant No. 192 25 00054); Human Robot Collaborative AI under its AME Programmatic Funding Scheme (Project No. A18A2b0046); National Research Foundation Singapore under its AI Singapore Programme (Award Grant No: AISG-100E-2018-006), and by A*STAR under its RIE2020 Advanced Manufacturing and Engineering Domain (AME) Programmatic Grant (Grant No. A1687b0033, Project Title: Spiking Neural Networks).

Approval of the version of the manuscript to be published.